\documentclass[10pt, twocolumn, letterpaper]{article}
\usepackage{subcaption} 
\usepackage[usenames, dvipsnames, svgnames, table]{xcolor}
\usepackage{wacv}
\usepackage{times}
\usepackage{epsfig}
\usepackage{graphicx}
\usepackage{amsmath}
\usepackage{amssymb}

\makeatletter
\@namedef{ver@everyshi.sty}{}
\makeatother

\usepackage{braket}
\usepackage{algorithm} 
\usepackage{algpseudocode} 
 
\algrenewcommand\algorithmiccomment[1]{\hfill\textit{ #1 }}

\DeclareMathOperator*{\argmin}{argmin}

\newcommand{\first}[1]{\textcolor{red}{#1}}
\newcommand{\second}[1]{\textcolor{blue}{#1}}
\newcommand{\third}[1]{\textcolor{ForestGreen}{#1}}

\newcommand{\bc}{\mathbf{c}}
\newcommand{\bx}{\mathbf{x}}
\newcommand{\bh}{\mathbf{h}}
\newcommand{\model}{\mathcal{M}}
\newcommand{\normal}{\mathcal{N}}
\newcommand{\patch}{\mathcal{P}}
\newcommand{\object}{\mathcal{O}}

\newcommand{\bb}[1]{\boldsymbol{\mathbf{#1}}}

\usepackage[pagebackref=true,
			breaklinks=true,
			letterpaper=true,
			colorlinks,
			bookmarks=false]{hyperref}
			
\newcommand{\labelphantom}[1]{%
  \parbox{0pt}{\phantomsubcaption\label{#1}}%
}

\usepackage{multirow}

\wacvfinalcopy 

\def\wacvPaperID{938} 

\ifwacvfinal\fi
\begin{document}

\title{Part-based Tracking by Sampling}

\author{{\centering George De Ath \hspace{0.5cm} Richard M. Everson} \\
		University of Exeter, United Kingdom\\
		{\tt\small \{g.de.ath,r.m.everson\}@exeter.ac.uk}
}

\maketitle
\ifwacvfinal\thispagestyle{empty}\fi

\begin{abstract}
We propose a novel part-based method for tracking an arbitrary object in 
challenging video sequences. The colour distribution of tracked image patches
on the target object are represented by pairs of RGB samples and counts of how
many pixels in the patch are similar to them. Patches are placed by segmenting
the object in the given bounding box and placing patches in homogeneous regions
of the object. These are located in subsequent image frames by applying 
non-shearing affine transformations to the patches' previous locations, locally
optimising the best of these, and evaluating their quality using a modified
Bhattacharyya distance. In experiments carried out on VOT2018 and OTB100
benchmarks, the tracker achieves higher performance than all other part-based
trackers. An ablation study is used to reveal the effectiveness of each
tracking component, with largest performance gains found when using the patch
placement scheme.
\end{abstract}

\section{Introduction}
\label{sec:introduction}
The tracking of arbitrary objects, also know as model-free tracking, has
applications across many different fields including video surveillance, 
activity analysis, robot vision, and human-computer interfaces. The goal of
model-free tracking is to determine the location of an unknown object, 
specified only by a bounding box in the first frame of a sequence, in all
subsequent frames. Although the topic has attracted much recent research 
interest and great strides have been made in tracking performance in recent 
years \cite{VOT2018}, visual object tracking still remains a challenging 
problem because of the many difficult real-world tracking scenarios that can 
be encountered. These include camera motion, illumination change, object motion
and size change, as well as occlusion and self-occlusion \cite{VOT2013}.

The way an object is represented, \ie its model, is arguably the most important
component in a tracking algorithm. Objects can either be represented in a 
holistic manner \cite{LADCF,MFT,LSART,STRUCK,KCF,MDNET,SRDCF,CCOT}, with a 
part-based representation, \cite{BST,DPCF,elastic_patches,SHCT,GGTv2,DPT,DGT,
WPCL,AIHSS,BHMC,BDF,MATRIOSKA}, or as a combination of the two 
\cite{LGT,CDTT,ANT}. Holistic methods represent the object using one global 
model that characterises the entire region that the object resides in, 
typically its bounding box. However, if large amounts of deformation or 
occlusion occur then global models can fail to robustly track the target
\cite{GGTv2}. A promising way of countering these types of problems is through
part-based methods, which use smaller, localised models to represent 
sub-regions of the object, known as \textit{parts} or \textit{patches}, which
together can then be used to estimate the object's overall location.

Here we construct a novel part model of the tracked object, drawing inspiration
from ViBe \cite{VIBE}, a non-parametric pixel-based algorithm that is one of 
the simplest and most effective background subtraction techniques
\cite{VIBE_review}. It models each pixel in a image by storing samples of their
values over time, and using the set of these samples to model the pixel's color
distribution. We reverse the background subtraction problem and model the 
image's foreground (\ie the object) rather than its background. The pixel-based
representation is extended to a part-based one by using samples of features
(usually colours) taken from image patches to characterise the distribution of
a part's features. We also develop an object localisation scheme to find the
modelled object's parts in subsequent frames, along with an update scheme to
update the models once their locations have been predicted. 

A commonly overlooked areas of model-free tracking is the problem of
determining which pixels, within the given bounding box, actually correspond
to the object \cite{init_paper}. For trackers that use a global model, such as
correlation filters, this is less of a concern because the majority of the
region given to the tracker (typically over 70\% \cite{VOT2016_seg}) belongs to
the object, resulting in a small proportion of spurious pixels in their
model(s). However, for part-based trackers, the placement of their parts is of
paramount importance because if a part is initialised on the background of the
object then this part will continue to match a stationary region in the image,
rather than the moving object. This can lead to some parts trying to follow the
object, while others remain static, ``tracking" part of the background,
resulting in a  reduction in tracking performance. We tackle this using a novel
object part placement scheme.

In summary, we propose a part-based visual tracking framework which has the 
following novel contributions:
\begin{itemize}
\itemsep=-2pt
\item A sparse, part-based visual model that empirically characterises the 
underlying colour distribution of the image patches with clusters in colour
space. 
\item An object part placement mechanism that, given a bounding box 
containing an object, segments the object and selects part locations in
homogeneous regions on the object.
\item An object localisation method that models an object's motion 
using global non-shearing affine transformations followed by localised patch 
location optimisation, mirroring typical object movement.
\end{itemize}

The remainder of the paper is organized as follows: \S\ref{sec:related_work}
provides a review of related methods, \S\ref{sec:PBTS} describes our tracking
method in detail, and in \S\ref{sec:experiments} we present experimental
results on the VOT2018 and OTB100 datasets, including an ablation study to
highlight the contribution of each component of the tracker. Conclusions are
drawn in \S\ref{sec:conclusion}.

\section{Related Work}
\label{sec:related_work}
We review the main components of part-based tracking methods, focussing on how
parts are placed on objects, the way in which parts are represented and 
inter-part constraints. General surveys of visual object tracking methods may
be found in \cite{vot_perf_measures, CF_survey, img_proc_for_tracking, ALOV,
app_model_survey, sparse_coding_review, adaptive_modelling_review,
advances_in_vt, motion_capture_survey, object_tracking_survey,
real_time_obj_track_survey}.

\subsection{Object Part Placement}
\label{sec:related_work:part_placement}
Part-based methods usually select the locations of the patches representing
parts in an \textit{ad hoc} fashion, either distributing them uniformly across
the object, or trying to select regions that have good properties in relation
to their part model. Some place their constituent parts uniformly on a grid
across the bounding box \cite{LGT,DCCO,elastic_patches}, while others place 
patches so that they touch each other and cover the entire bounding box 
\cite{DPCF,RTCF,DPT,LGCF,fragtrack}. Those that place their patches in this 
fashion make no assumptions as to which pixels within the bounding box belong 
to the object.

Other techniques place their parts based on the type of part-model they have.
The ANT tracker \cite{ANT}, for example, chooses patch locations that maximise
the chance of having good optical flow characteristics. Methods that represent
object parts as superpixels \cite{DGT, SHCT, GGTv2} select superpixels that lie
wholly within the bounding box (\ie have no pixels overlapping it). However,
\cite{CDTT} also selecting superpixels that overlap the bounding box, because
since superpixeling segments an image into homogeneous regions, any superpixels
that contain background are likely to contain only background pixels, meaning
that they can quickly prune the poorly matching superpixels later in tracking.

PBTS uses superpixels to segment the initial bounding box, but we first label
pixels as object or background, using only the initial image and the bounding
box \cite{init_paper}. We then superpixel the pixels in the bounding box that
have a high likelihood of belonging to the object, thereby drastically reducing
the chance of placing patches on pixels that do not belong to the object.

\subsection{Object/Part Representation and Matching}
\label{sec:objectp-repr-match}
Parts are generally represented using  either keypoints, correlation filters,
or histograms of features extracted from the parts. Keypoint methods represent
an object at a set of distinguished feature points. Keypoints are characterised
by the local image texture using, \eg, ORB \cite{ORB}, SIFT \cite{SIFT}, or
SURF \cite{SURF}. The approximate invariance to image scale and rotation of
these descriptors   allows them to be matched to the corresponding points in
the subsequent frame.

Correlation filters match regions in successive frames by maximising the
correlation between an image patch in two frames. They are effective and
efficient in object tracking \cite{VOT2018}, but the range over which the
correlation is maximised is limited to half the width or height of the template
being matched \cite{DPT}, limiting their use for matching parts, which are
generally small compared to the object movement.

Histograms capture the feature distribution of the patch corresponding to a
part. Commonly used features are the RGB pixel intensities, but they have been
used widely with a variety of features, \eg \cite{LGT, DPT, BHMC}. Patch
similarity, expressing the overlap between the probability distributions 
estimated by the histograms, can be computed in a variety of ways, methods
based on the Bhattacharyya distance \cite{mean_shift, bhattacharyya_original}
being popular. Here we use a novel object patch representation and compare the
similarity of patches using a modified version of the Bhattacharyya distance.

\begin{figure*}[t] 
\centering
\includegraphics[width=\textwidth, clip, trim={14 157 25 5}]{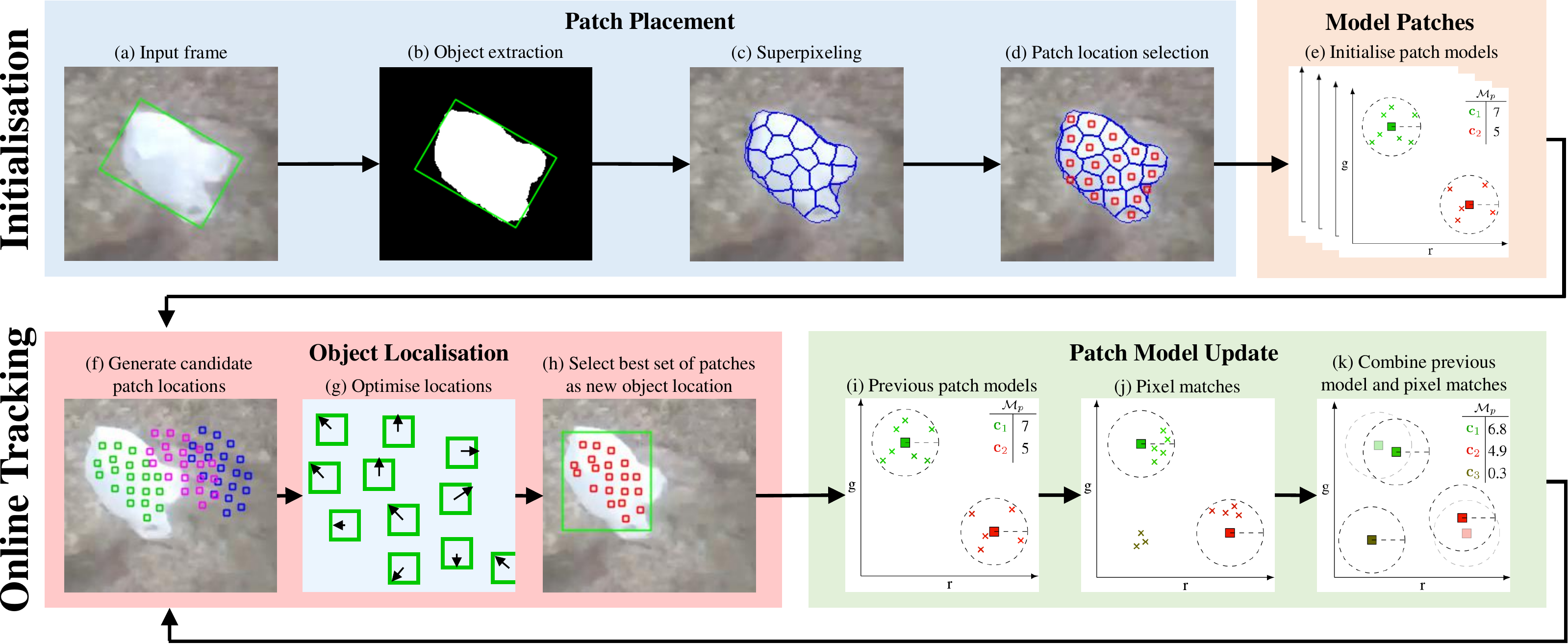}
\labelphantom{fig:tracking_pipeline:a}
\labelphantom{fig:tracking_pipeline:b}
\labelphantom{fig:tracking_pipeline:c}
\labelphantom{fig:tracking_pipeline:d}
\labelphantom{fig:tracking_pipeline:e}
\labelphantom{fig:tracking_pipeline:f}
\labelphantom{fig:tracking_pipeline:g}
\labelphantom{fig:tracking_pipeline:h}
\labelphantom{fig:tracking_pipeline:i}
\labelphantom{fig:tracking_pipeline:j}
\labelphantom{fig:tracking_pipeline:k}
\caption[]{Schematic diagram of PBTS. For ease of illustration in (e) and (i-k)
pixels are represented as (r,g) pairs rather than (r,g,b) triples.}
\label{fig:tracking_pipeline}
\end{figure*}

\subsection{Part Constraints and Localisation}
\label{sec:part-constr-local}
The way in which part-based methods relate their parts to one another 
geometrically plays an important role in their tracking performance. Some
enforce no explicit geometric relationships \cite{BHMC,CDTT}, others a 
star-based topology \cite{DGT,TSIT,MATRIOSKA,WPCL}, and some have a 
fully-connected sets of parts \cite{AIHSS,DPT}. Intermediate between these are
local constraints between neighbouring parts, \eg \cite{SHCT, LGT, GGTv2}.

Implicit constraints on a patch's location may also result from limiting the
range of potential patch locations searched  in subsequent frames, often by
modelling part motion between frames by global affine or rigid transformation,
after which part locations are individually optimised in a restricted region
around the global transformation \cite{elastic_patches, LGT}.

In PBTS we also implicitly constrain the potential locations patches to be
evaluated, based on our observation that an object's motion between frames is
largely rigid and only varies slightly from this rigidity locally. We therefore
generate a set of candidate global non-shearing affine transformations and
apply them to each part's location in the previous frame. The best of these is
then used as the base transformation for individual optimisations of each
part's location in a small window around the best globally transformed
location, thus allowing a localised non-rigidity.

\section{Part-based Tracking by Sampling}
\label{sec:PBTS}
Fig.~\ref{fig:tracking_pipeline} gives an overview of PBTS, which we first
describe in outline before giving details about each stage.

The tracker is initialised in the first frame of a video sequence from the
supplied bounding box containing the object to be tracked 
(Fig.~\ref{fig:tracking_pipeline:a}). As described in 
\S\ref{sec:patch-placement}, pixels likely to belong to the object in the
bounding box are identified (Fig.~\ref{fig:tracking_pipeline:b}) and
superpixelled (Fig.~\ref{fig:tracking_pipeline:c}), and a patch/part chosen at
the centre of each superpixel (red squares in 
Fig.~\ref{fig:tracking_pipeline:d}). Patches are represented by their centres 
$\bx_p$ and a model $\model_p$ of the colour distribution of pixels in the
patch. As described in \S\ref{sec:object-model}, $\model_p$ comprises of
samples from the patch's pixels together with counts describing the number of
similar pixels in the patch, Fig.~\ref{fig:tracking_pipeline:e}. In subsequent
frames the patches are transformed from their locations in the previous frame
by randomly chosen global non-shearing linear transformations to candidate
locations; represented by green, magenta and blue squares in 
Fig.~\ref{fig:tracking_pipeline:f}. The most likely of these is determined by
matching the colour models of patches in the previous frame to their
transformed locations, after which the individual patch locations are locally
optimised (Fig.~\ref{fig:tracking_pipeline:g}). Finally, the patch colour
models are updated (\S\ref{sec:model-update}). In outline, the sparse colour
models are updated by moving the features in the old colour model towards the
matching pixels in the new patch, and introducing new features to represent
previously unseen pixel values or pruning now redundant features
(Figs.~\ref{fig:tracking_pipeline:i} -- \ref{fig:tracking_pipeline:k}).

\subsection{Object Model}
\label{sec:object-model}
We represent an object $\object$ as an ordered set of $P$ patches
$\object = \set{\patch_p}_{p=1}^P$. Patches are rectangular with width and
height of $w_x$ and $w_y$ pixels respectively, and each patch
$\patch_p = (\bx_p, \model_p)$ is characterised by its location $\bx_p$ and a
colour model $\mathcal{M}_p = \set{ (\bc_1, h_1), ..., (\bc_{S}, h_{S}) }$,
containing $S$ pairs of centres $\bc_s$ and their corresponding match counts
$h_s$; see Fig.~\ref{fig:tracking_pipeline:e}. Centres are $(r, g, b)$ samples
from pixels in the patch and the counts $h_s$ indicate how many pixels within
the patch match the centre $\bc_s$. A pixel $\bb{I}_i = (r_i, g_i, b_i)$ is
said to match the centre if $\lVert \bb{I}_i -\bc_s \rVert < R$ for radius $R$.
Thus the colour model can be thought of a sparse empirical histogram of the
frequency of colours in the patch. Although we describe the colour model in
terms of R, G, B colour descriptors the method naturally generalises to other
colour spaces or more sophisticated colour or texture descriptors.

The colour model is initialised from the pixels $\bb{I}_i$ in the patch region,
which we denote by $\Omega_p$. We start by randomly selecting a pixel 
$\bb{I}_i$ as the first sample in the model $\bc_1 = \bb{I}_i$, with match
count $h_1 = 1$. Successive pixels are randomly selected, without replacement,
and compared to the centres in the model. If the pixel matches, the count of
the centre it is closest to is incremented by one: $h_{s'} := h_{s'} + 1$ where
$s' = \argmin_{s'} \lVert\bb{I}_i - \bc_{s'}\rVert$. If it does not match any
centre (\ie $\lVert\bb{I}_i - \bc_s\rVert > R \,\forall s = 1,\ldots,S$) then
that pixel is added to the model as a new centre: $\bc_{S+1} = \bb{I}_i$ with
$h_{S+1} = 1$. This process is repeated until all pixels in the patch have been
evaluated once. Finally, if there are more than a predetermined number 
centre-count pairs $S_{max}$, then the pairs with the lowest counts are removed
until the model only contains $S_{max}$ pairs of centres and counts.

This initialisation yields a kind of pseudo-clustering, without the need for
an explicit clustering algorithm, tailored to producing clusters of a fixed
radius ($R$), centred on pixel samples. This stochastic sampling method is much
less computationally intensive than traditional clustering techniques. Since
the models are initialised from relatively homogeneous patches and samples are
therefore close in colour space, it  yields similar cluster centres regardless
of the sampling sequence. Limiting the number of centres to $S_{max}$ in each
model effectively reduces model over-fitting by excluding rare unrepresentative
samples.

Unlike histograms that partition colour space into rectangular bins with
predetermined centres on a rectangular grid, this model makes no assumptions as
to how the colour space should be partitioned. Instead, the use of colour
samples themselves as bins yields a sparse representation as, unlike
conventional histograms, there are no empty bins.

\subsection{Patch Similarity}
\label{sec:patch-similarity}
Fundamental to PBTS is evaluating how well a candidate patch $\tilde{\patch}$
in the current frame matches $\patch$ without building a new model for the
candidate patch. The quality of the match is evaluated by comparing how well
pixels in $\tilde{\patch}$ match $\mathcal{M}_p$. This is achieved by counting
the number of pixels in $\tilde{\patch}$ that are within radius $R$ of each
centre $\bc_s $ in $\model_p$.\footnote{If a pixel matches more than one centre
in $\mathcal{M}_p$, (because two centres lie within $2R$ of one another) the
centre closest to the pixel is matched.} Define $\tilde{\bh}$ to be the vector
of match counts for pixels in $\tilde{\patch}$ to the $S$ centres of $\patch_p$
and let $\bh$ be the normalised vector of counts for $\patch$ itself. Both
vectors are normalised by dividing by the number of pixels in the patch. Then
the similarity between $\patch$ and $\tilde{\patch}$ can be quantified using a
modification of the Bhattacharyya distance \cite{mean_shift}. The Bhattacharyya
coefficient \cite{bhattacharyya_original} between two distributions, 
$\bb{p}$ and $\bb{q}$, where ${p}_j \geq 0$, ${q}_j \geq 0$,
$\sum {p}_j = \sum {q}_j = 1$ is given by:
\begin{equation}
BC\left( \bb{p}, \bb{q} \right) = \sum_{j=1}^{S} \sqrt{p_j q_j} \,.
\label{eqn:ch3:BC}
\end{equation}
Note that $0 \leq BC\left( \bb{p}, \bb{q} \right) \leq 1$ measures the overlap 
between two distributions, $\bb{p}$ and $\bb{q}$, with 
$BC\left( \bb{p}, \bb{q} \right) = 0$ iff there is no overlap. We define the 
modified version of the Bhattacharyya Distance (MBD) to be
\begin{equation}
MBD\left( \bb{p}, \bb{q} \right) = \left[ 1 - BC\left( \bb{p}, \bb{q} \right) \right]^b
\label{eqn:ch3:MBD}
\end{equation}
where $b \geq 0$ controls the weight given to good matches. A larger value of 
$b$ favours good matches, whereas they are down-weighted with smaller values. 
Setting $b = \tfrac{1}{2}$ recovers the original Bhattacharyya distance
\cite{mean_shift}, which is also equal to the Hellinger distance
\cite{hellinger_distance}. 

Using the modified Bhattacharyya distance we define the quality of the match
between $\model$ and $\tilde{\patch}$ as 
\begin{equation}
Q( \patch, \tilde{\patch}) = 1 - MBD( \bh, \tilde{\bh}).
\label{eqn:ch3:match_quality}
\end{equation}
The overall quality of a set of a candidate parts  
$\tilde{\object} = \set{\tilde{\patch}_p}_{p=1}^P$ representing an object is
thus the average similarity of the individual patches to the object being
matched:
\begin{equation}
L ( \object, \tilde{\object}) = \frac{1}{P}
\sum^P_{p=1} Q ( \patch_p, \tilde{\patch_p}).
\label{eqn:ch3:overall_match_quality}
\end{equation}
Note that after normalization by the number of pixels in a patch, the vector of
match counts $\sum \tilde{h}_j \leq 1$ because only the pixels that match each
centre in $\model_p$ are counted. This is a desirable property as it will
implicitly down-weight candidate patches that have fewer pixels matching the
model.

\subsection{Localisation}
\label{sec:localisation}
Empirically we observe that object movement between consecutive frames is
largely rigid and  it only has minor local deviations from rigidity
\cite{my_thesis}. More precisely, we find that object movement is well
represented by a rotation, an isotropic scaling and a translation. We therefore
match patches in a frame to those in the previous frame in two steps: we first
generate non-shearing affine transformations of the previous frame's patch
locations to match the rigidity assumption, and then optimise each patch's
location within a small window around the best rigid transformation to allow 
for local non-rigidity; see 
Figs.~\ref{fig:tracking_pipeline:f}~and~\ref{fig:tracking_pipeline:g}.

$G$ non-shearing affine transformations 
$\bb{A}_g  = \bb{T}(x,y) \bb{S}(s)\bb{R}(r)\bb{O}$ are generated as the
products of a translation to the origin $\bb{O}$, a rotation $\bb{R}(r)$ by $r$
radians, an isotropic scaling $\bb{S}(s)$ by a factor $s$ and a translation 
$\bb{T}(x,y)$ by $[x, y]^T$.  The parameters are drawn from random
distributions centred on the patch location in the previous frame, thus
$r \sim \normal(0, \sigma_r)$ and $s \sim \normal(1, \sigma_s)$ where
$\sigma_r = \pi/16$ and $\sigma_s = 0.02$ were chosen to match the typical
rotations and scale changes found in an extensive survey of the VOT2016 data
\cite{VOT2016_seg, my_thesis}. We observe that the distributions of inter-frame
vertical and horizontal translations are heavy-tailed and therefore draw
translation parameters from Laplace distributions centred on the object
location in the previous frame and with scale parameters
$w'\sigma_x$ and $h'\sigma_y$ where $w'$ and $h'$ are the predicted width
and height of the object in the previous frame and $\sigma_x = 0.15$ and
$\sigma_y = 0.1$ were determined empirically \cite{my_thesis}. Scaling the
translation distributions' length-scales by the object's predicted width and
height allows for the predicted movement of the object to be adjusted relative
to its size, because generally, larger objects move further than smaller
objects, but are comparatively similar when this movement is considered as a
proportion of their own size.

The patch locations resulting from the best $L$ candidates of the $G$ randomly
generated transformations are each locally optimised by an exhaustive search of
potential patch locations within in a square window with side length $W$
pixels, centred on the transformed patch at $\tilde{\bx}_p = \bb{A}_g\bx_p$.
The match quality of the patch centred at each location within the window is
evaluated and the location with the highest quality is selected as the patch's
new predicted location; Fig.~\ref{fig:tracking_pipeline:g}. If there are
multiple locations with the same match quality then the one closest to the
globally transformed location $\tilde{\bx}_p$ is selected, with equidistant
ties broken randomly.

Following preliminary experiments, we report the predicted bounding box of the
object as the axis-aligned bounding box (AABB) $20\%$ wider and taller than the
AABB that minimally encloses the patches after matching; 
Fig.~\ref{fig:tracking_pipeline:h}.

We note that while the methods of \cite{elastic_patches} and \cite{LGT} share
some similarities with our object localisation scheme, there are several
important differences. In contrast to \cite{elastic_patches}, who, after
applying an affine transformation to the set of patches, randomly move each
patch and evaluate its quality, we locally optimise within a small region 
around the affine transformed position. \cite{elastic_patches} also limit the 
class of transformations to only include translation, whereas we also include
both isotropic scaling and rotation. The localisation scheme of \cite{LGT} uses
the cross-entropy method to first find the optimal affine transformation and
then locally optimises each patch. Our methods differs from this as we locally
optimise the $L$ sets of patch locations with the highest quality, each set of
which can have different patch locations to other sets.

\cite{LGT} start a search for the optimal affine transform by sampling from a 
Gaussian distribution with a covariance of $20\bb{I}$ for all sizes of objects, 
similar to \cite{elastic_patches} who also sample from a Gaussian distribution 
with a scale parameter of $8$ pixels for both horizontal and vertical
translation. In contrast, we sample translations drawn from distributions whose
scale parameter is relative to the respective width or height of the object, as
we have observed that the size of inter-frame object motion is approximately
proportional to the object's size. 

\subsection{Model Update}
\label{sec:model-update}

In common with other models that represent the colour distribution with 
histograms such as \cite{fragtrack, ANT, BHMC}, we linearly interpolate the
probability of finding a colour in a patch between the original model and
the newly matched image region. Our sparse representation of the colour 
histogram means that in addition to updating the centre-counts pairs
$(\bc_i, h_i)$, we need to be able to introduce new and remove redundant
centre-count pairs. 

Let $\Omega_p$ be the set of pixels comprising the image region that matches
model $\model_p$ and let
$\omega_{p,s} = \set{\bb{I} \in \Omega_p | \lVert\bb{I} - \bc_s\rVert <  R}$
be the set of pixels in $\Omega_p$ which match the $s$-th centre in the model
(\ie pixels inside the dashed circles in Fig.~\ref{fig:tracking_pipeline:j}).

Then the model counts are updated by linearly interpolating:
\begin{equation}
  \label{eq:update-counts}
  h_{p,s} := \beta_c |\omega_{p,s}|  + (1-\beta_c) h_{p,s} 
\end{equation}
where $\beta_c$ is an update rate. Similarly, the locations of the centres are
updated to move them towards the mean of the matched pixels:
\begin{equation}
  \label{eq:update-centre-locations}
  \bc_s := \beta_s \frac{1}{|\omega_{p,s}|}\sum_{\bb{I}_j\in
    \omega_{p,s}}\bb{I}_j + (1-\beta_s) \bc_s
\end{equation}
The use of two update rates, $\beta_c$ and $\beta_s$, allows for the centres
and counts to be adapted at different speeds. This is an important property as
when an image region changes colour due to illumination intensity the
brightness of each colour in the region increases, rather than the relative
proportion of colours. Updating $\bc_j$ has the effect of moving the centre of
the sphere containing matching pixels towards (or past if $\beta_s > 1$) the
matches' centre of mass. It allows the model to \textit{follow} the region of
colour space that it characterises as it changes over time. In the standard
histogram-based approach, when the centres change over time they move from
matching one bin to matching another. When this occurs there will be a sudden
loss of probability mass within the histogram, resulting in poorer matching.
Following extensive experimentation we found that $\beta_s = 1.7$ yields the
best tracking performance, meaning that the update is predicting where the
colour distribution of the patch will lie in the subsequent frame.

New colour centres corresponding to pixels $\Omega^-_p$ which do not match the
original colour model (\ie pixels outside the dashed circles in 
Fig.~\ref{fig:tracking_pipeline:j}) are incorporated in $\model_p$ by first
creating a new patch model $\mathcal{M}_p^-$  from them. Each of the counts
$h_s^-$ is scaled by $\beta_c $ and the centre-count pairs in $\model_p^-$
are added to $\model_p$; \eg $(\bc_3, h_3)$ in 
Fig.~\ref{fig:tracking_pipeline:k}. The scaling by $\beta_c $ is done in order
to match the behaviour of count updating, as this introduces the centre-count
pair as though they were already in the model with a previous count of $0$; 
cf. Eq.~\eqref{eq:update-counts}.

Lastly, the centre-count pairs with $h_s < \beta_c$ are removed from 
$\model_p$. This allows for centres that have not been seen recently to be
removed in order to limit the model's size and computational complexity. A
threshold of $\beta_c$ was chosen as it is the smallest value that the count
for a new centre-count pair can have if it were added to the model at the same
time-step, \ie it had a value of $h_s = 1$ before scaling.

\subsection{Patch Placement}
\label{sec:patch-placement}
Since our colour model provides a compact representation of the colours in a
patch, we seek to select spatially compact regions of homogeneous colour as
patches. This is achieved by superpixelling \cite{superpixels} the supplied
bounding box; see Fig.~\ref{fig:tracking_pipeline:g}. Superpixelling 
over-segments images into perceptually meaningful regions that are generally
uniform in colour and texture. As superpixels tend to adhere to colour and
shape boundaries, they also retain the image's structure
\cite{superpixel_review}. As illustrated in Fig.~\ref{fig:tracking_pipeline:d},
object parts are placed at the centre of superpixels that reside within the
object's predicted location as determined by the alpha-matting segmentation
scheme described in \cite{init_paper}.

A parameterless version (SLICO\footnote{\url{https://ivrl.epfl.ch/research-2/research-current/research-superpixels/\#SLICO}})
of the SLIC superpixelling algorithm \cite{SLIC} is used to segment the region
surrounding the predicted location of the object into approximately $P$
superpixels. The zero-parameter version of the state-of-the-art SLIC 
\cite{superpixel_review} allows for each superpixel to have its own compactness
parameter, so that regular shaped superpixels are generated for both smooth and
rugged image regions.

The location of the first patch is taken as the centroid of the largest
superpixel. Successive patches are then greedily initialised to the centroid of
the largest unassigned superpixel if doing so would cause the patch to overlap
with all other patches with a proportion of its area less than $\gamma$. This
process is repeated until either $P$ patch locations have been selected or
there are no more superpixels left to consider.

The degree of permitted overlap $\gamma$ between patches controls the density
of the patches across the object. For larger objects, where patches do not
reach the boundaries of superpixels, its value is immaterial. However, for
smaller objects, patches may overlap multiple superpixels, if the area of the
superpixels is less than that of the patches' area, or when the superpixels are
roughly rectangular with dimensions shorter or longer than the patches.
Limiting the patch overlap in these cases reduces the amount redundant
information in neighbouring patches.

An alternative would be to initialise patches on the boundaries between
superpixels where there tends to be high contrast so that patches would be
centred on image features such as corners or ``keypoints''. However,
experiments show that initialisation in homogeneous regions results in better
tracking performance.

\section{Experiments}
\label{sec:experiments}
The tracker was evaluated on the VOT2018 benchmark \cite{VOT2018}, which 
provides a tracking dataset with fully annotated frames, and reports the 
performance of a large number of state-of-the-art trackers. The dataset
comprises of 60 sequences, containing difficult tracking scenarios such as
occlusion, scale variation, camera motion, object motion change, and
illumination changes. We followed the VOT challenge protocol, with the tracker
initialised on the first frame of a sequence using the ground-truth bounding 
box provided, and reinitialised if the tracker drifted away from the target. 
Trackers were evaluated in terms of their accuracy (target localisation),
robustness (failure frequency), and Expected Average Overlap (EAO). For full
details see \cite{VOT2018}. PBTS was compared to the 10 best part-based
trackers in VOT2018 as well as the top 3 state-of-the-art trackers.

The tracker was also evaluated on the OTB100 benchmark \cite{OTB100} which
can be viewed as complementary to the VOT evaluation as its main focus is on
unsupervised tracking. However, trackers which explicitly include re-detection
mechanisms are at an advantage when compared to the those trackers evaluated on
the VOT benchmark, which generally do not include any long-term tracking
components. We compare the performance of PBTS to the top-performing trackers
on 75 colours sequence in OTB100 since PBTS uses only colour information 
(RGB features). PBTS is compared with other trackers using their published
results\footnote{\url{http://cvlab.hanyang.ac.kr/tracker_benchmark/}} on the
same 75 colour videos. Performance is measured using success and precision
plots, which are measure-threshold plots of the proportion of tracked frames
that have an intersection-over-union (IOU) overlap greater than some threshold
and the proportion of frames that have a centre error (the distance between the
centroids of the predicted and ground-truth bounding boxes) of less than a
threshold.

We also report an ablation study which highlights the contribution of each
component of the tracker, together with a qualitative analysis of success and
failure modes.

\subsection{Implementation Details}
\label{sec:implementation_details}
PBTS was implemented in Python 3.6 on an Intel i5-4690 CPU with 32Gb RAM, and
runs at $\approx 15$ frames per second, calculated on the VOT2018 sequences
with no GPU or multi-threading. Its parameters were chosen via extensive 
cross-validation experimentation \cite{my_thesis}.

\noindent\textbf{Object Representation:}
Objects were modelled using $P = 35$ patches of size $w_x = w_y= 5$ pixels,
with the colour model using RGB pixel features. A matching distance of $R=20$
\cite{VIBE} was used, with a MBD coefficient value of $b = 1.4$ used for 
up-weighting better matching patches. Patch update parameters were 
$\beta_c = 0.05$ and $\beta_s = 1.7$.

\noindent\textbf{Patch Placement and Model Initialisation:} 
Using the notation of \cite{init_paper}, we set the object segmentation 
parameters as $\rho^- = 0.8$, $\rho^+ = 1.2$, $\tau = 0.85$, and 
$\lambda = 10^{-2}$. Patches were placed on objects with a maximum patch area 
overlap of $\gamma = 0.25$, and the number of samples limited in each model to 
$S_{max} = 10$ during initialisation. Performance was found to be roughly 
constant for $S_{max} \in [8,20]$.

\noindent\textbf{Object Localisation:}
Parameters for the random generation of non-shearing affine transformations are
given in \S\ref{sec:localisation}. $G=1000$ transforms were sampled, with the
best $L=100$ locally optimised. We found no performance gain by increasing $L$
or $G$. During local optimisation a patch may move in any direction 2 pixels
per time-step ($W = 5$).

\subsection{Evaluation on the VOT2018 Dataset}
\label{sec:eval-vot2018-datas}
\begin{table}[t]
\centering
\resizebox{\columnwidth}{!}{%
\setlength\tabcolsep{2pt}
\begin{tabular}{lcccllccc}
\cline{1-4} \cline{6-9}
Tracker & EAO & AO & R && Tracker & EAO & AO & R \\ \cline{1-4} \cline{6-9}

\textbf{PBTS} & \first{0.196}  & 0.427          & \first{1.723} && BDF \cite{BDF} & 0.093          & 0.336          & 4.200\\

ANT \cite{ANT} & \second{0.168} & \third{0.439}  & \second{2.250} && $\text{Matflow}\dagger$ & 0.092          & 0.362          & 4.550\\

DPT \cite{DPT} & \third{0.158}  & \second{0.463} & \third{2.567} && FragTrack \cite{fragtrack} & 0.068          & 0.325          & 6.650\\

LGT \cite{LGT} & 0.144          & 0.403          & 2.641 && Matrioska \cite{MATRIOSKA} & 0.065          & 0.365          & 6.900 \\
\cline{6-9}
$\text{DFPReco}\dagger$ & 0.138          & \first{0.467}  & 2.983 && LADCF \cite{LADCF} & 0.389          & 0.507          & 0.567 \\

FoT \cite{FoT} & 0.130          & 0.385          & 3.667 && MFT \cite{MFT} & 0.385          & 0.496          & 0.500\\

BST \cite{BST} & 0.116          & 0.244          & 3.136 && SiamRPN \cite{SiamRPN} & 0.383          & 0.567          & 0.900 \\
\cline{1-4} \cline{6-9}
\end{tabular}%
}%
\vspace{0.1cm}%
\caption{Results of PBTS on the VOT2018 benchmark compared to part-based 
trackers and state-of-the-art trackers (lower-right). Note that $\dagger$ 
indicates that details of the tracker are only published in the VOT2018 
benchmark \cite{VOT2018}.}
\label{tbl:VOT2018_results}
\end{table}

Table~\ref{tbl:VOT2018_results} shows the results of PBTS compared with the top
10 part-based trackers and top 3 state-of-the-art trackers on the VOT2018
benchmark.\footnote{Results previously reported for PBTS in \cite{VOT2018} did
not include any centre-count updating ($\beta_c = 0$) and used a MBD weighting
of $b = {1}/{2}$.} Results for other trackers are taken from \cite{VOT2018}.
The average overlap (AO) and robustness (R) scores are averaged over each
video, and so are per-video scores. As can be seen, PBTS outperforms all other
part-based trackers in terms of both EAO and robustness, and places fourth in
terms of accuracy. Given the relative simplicity of the tracker, compared to
those using global models (\eg ANT \cite{ANT}, DPT \cite{DPT}, LGT \cite{LGT})
to guide tracking in addition to their part-based, local models, the PBTS
methodology performs well. However, compared to the current state-of-the-art
methods (those in the lower-right portion of the table) our tracker performs
considerably worse with respect to all three attributes. We suggest that this
is due to both the tracker architecture and the types of features used for
object representation. The 10 top-performing trackers on the benchmark
\cite{VOT2018} all use deep neural network features (combined with various
other hand-crafted features) in either correlation filter-based or CNN-based
frameworks.

\subsection{Evaluation on the OTB100 Dataset}
\label{sec:eval-otb100-datas}
\begin{figure}[t]
\newcommand{\ww}{0.239}%
\begin{subfigure}[t]{\ww\textwidth}
\includegraphics[width=\textwidth]{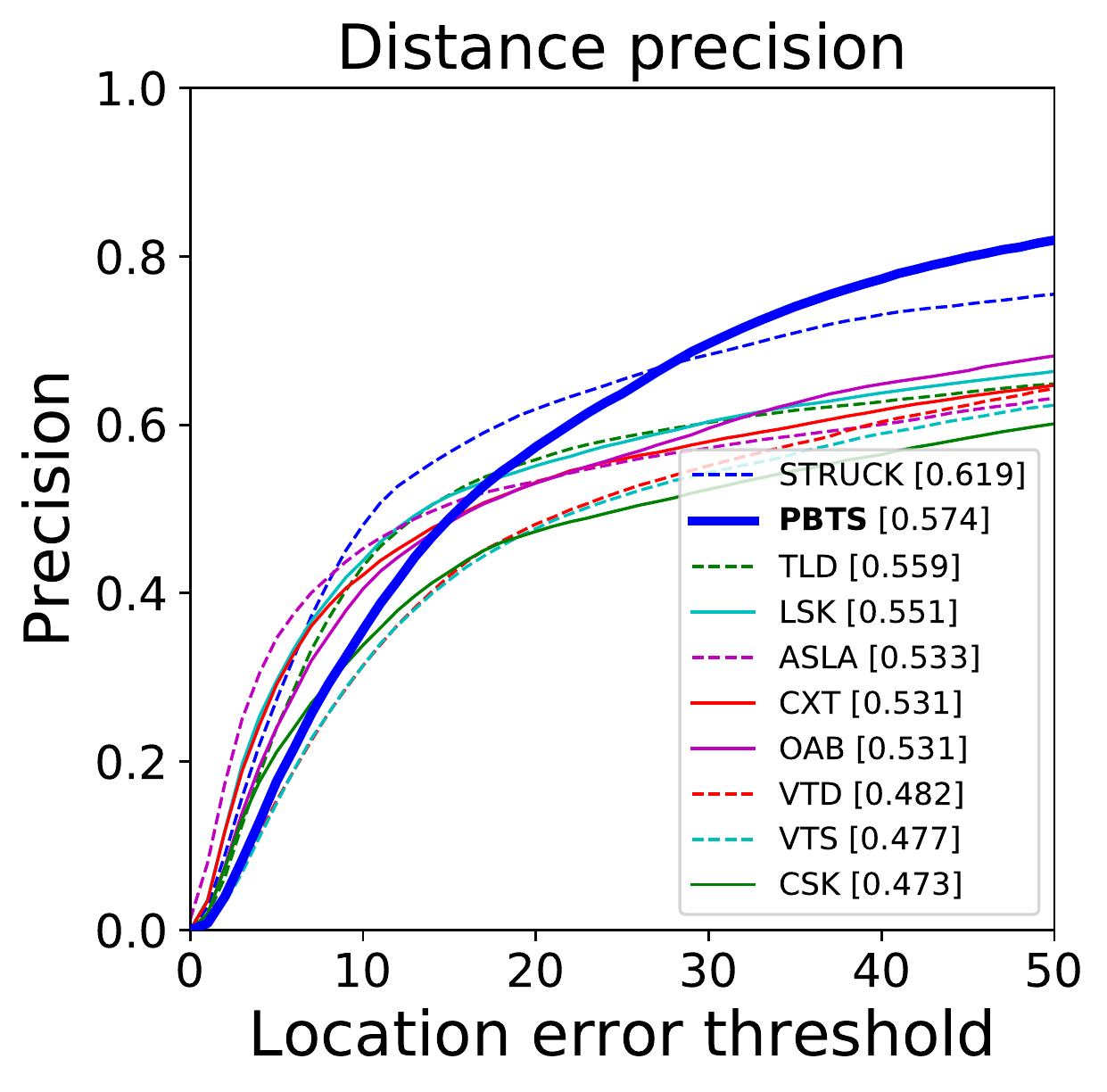}%
\end{subfigure}%
\begin{subfigure}[t]{\ww\textwidth}
\includegraphics[width=\textwidth]{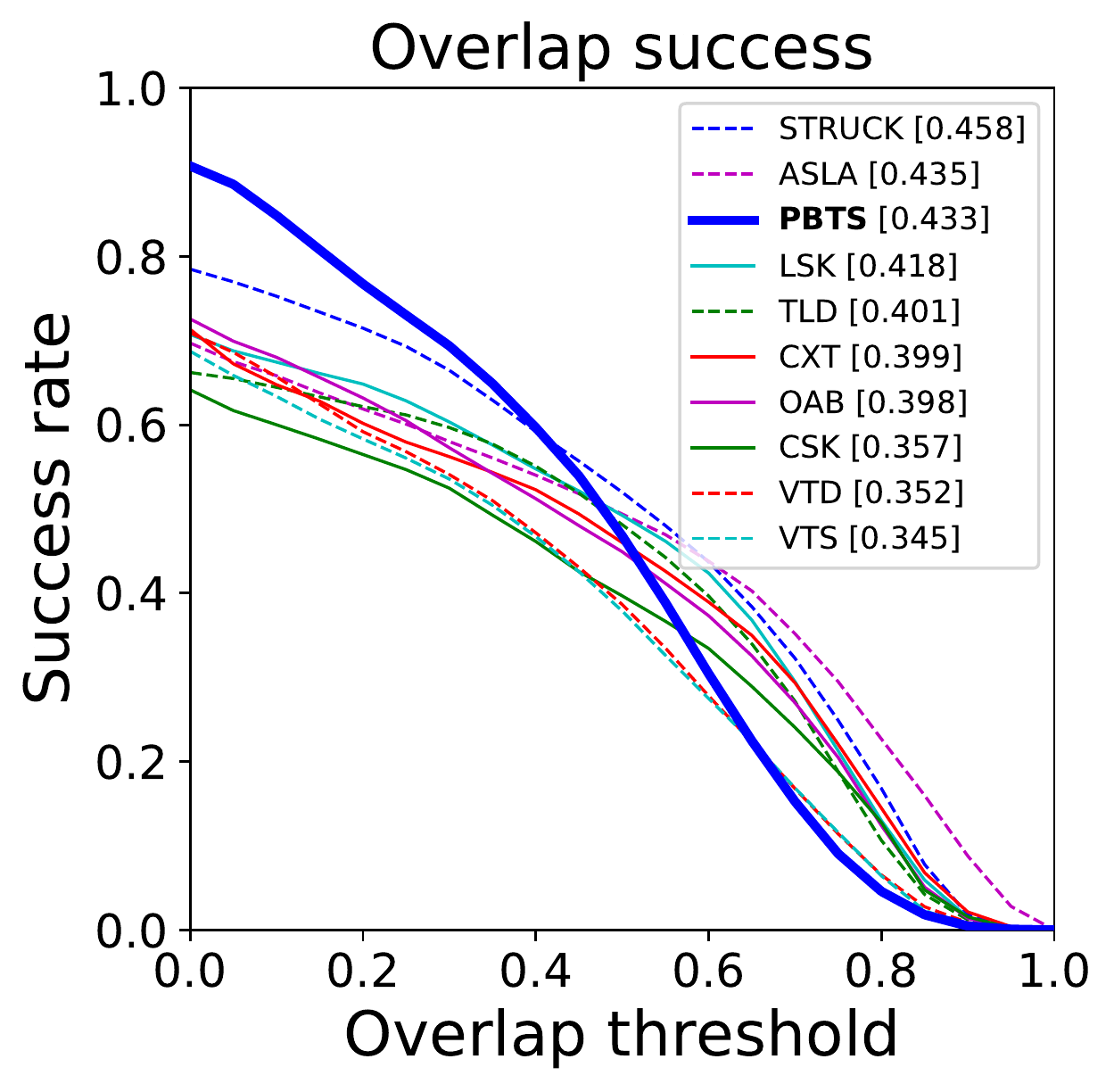}%
\end{subfigure}%
\caption{Precision and success plots using one-pass evaluation on the OTB100
dataset. The precision of the tracker with a 20 pixel threshold and the area
under the success plot are shown in brackets in the left and right-hand plots 
respectively. Better trackers have distance precision curves towards the 
top-left of the plots and overlap success curves towards the top-right of the 
plots.}
\label{fig:OTB100_colourvideos_overall}
\end{figure}

We compare PBTS to the top-performing trackers on the OTB100 OPE benchmark, 
namely STRUCK \cite{STRUCK}, CXT \cite{CXT}, ASLA \cite{ASLA}, TLD \cite{TLD},
CSK \cite{CSK}, LSK \cite{LSK}, VTD \cite{VTD}, OAB \cite{OAB}, and 
VTS \cite{VTS}. The performance of PBTS over all colour sequences is shown in
Fig.~\ref{fig:OTB100_colourvideos_overall}. Overall, PBTS performs well on the
benchmark, achieving second and third place in terms of precision and overlap
respectively. The general shape of the overlap success plots 
(Fig.~\ref{fig:OTB100_colourvideos_overall} (right)) of PBTS tends to be more
\textit{S}-shaped than the majority, which we suspect is due to the
underestimation of the bounding box; this is discussed further in 
\S\ref{sec:qualitative_analysis}. PBTS generally achieves a higher level of
precision than the other trackers once the location error threshold greater
than approximately 25 pixels. We believe that this is due to the robustness of
the part-based formulation of the tracker, as demonstrated by its performance
on the VOT2018 benchmark. Even if several object parts have drifted, the
majority will continue to track the object, despite the fact that the drifted
parts may be a significant distance from the main group of parts tracking the
object, leading to a good level of a precision at the expense of overlap. This
means that the predicted bounding box's centre may be away from the centre of
the object, but closer than typical holistic trackers (which model the entire
bounding box using one appearance model). In the holistic case, if the tracker
begins to model a region containing a large amount of background, the entire
tracker will tend to drift off the object, resulting in failure.

\subsection{Ablation Study}
\label{sec:ablation-study}
\begin{table}[t]
\centering
\setlength\tabcolsep{3pt}
\resizebox{\columnwidth}{!}{%
\begin{tabular}{lcccccc}
\hline
& \multirow{2}{*}{\textbf{PBTS}} & Default & No local     & No model & No Alpha & Uniform Patch \\
&      & MBD     & optimisation & update   & matting  & placement \\ \hline
EAO & \first{0.196} & \second{0.179}  & \third{0.169} & 0.162 & 0.127 & 0.118 \\
AO & \first{0.427} & 0.353 & \second{0.396} & \third{0.386} & 0.245 & 0.234 \\
R & 1.723 & \third{1.658} & 2.346 & 2.194 & \first{1.500} & \second{1.591} \\ \hline
\end{tabular}%
}%
\vspace{0.05cm}%
\caption{Ablation study using the VOT2018 benchmark.}
\label{tbl:ablation_results}
\end{table}

\begin{figure*}[t]
\centering
\newcommand{\ww}{0.099}%
\includegraphics[width=\ww\textwidth]{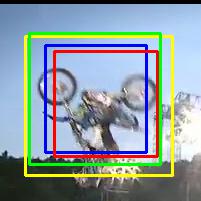}\hfill%
\includegraphics[width=\ww\textwidth]{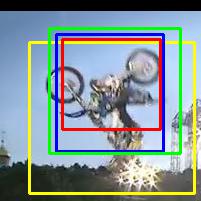}\hfill%
\includegraphics[width=\ww\textwidth]{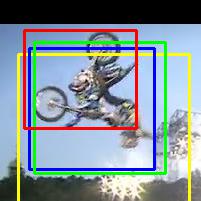}\hfill%
\includegraphics[width=\ww\textwidth]{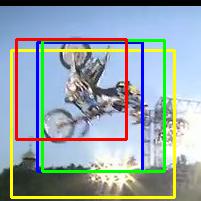}\hfill%
\includegraphics[width=\ww\textwidth]{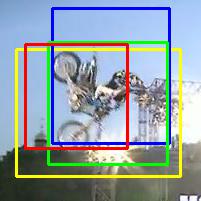}\hfill%
\includegraphics[width=\ww\textwidth]{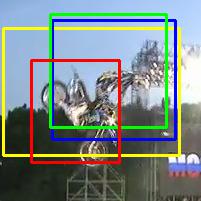}\hfill%
\includegraphics[width=\ww\textwidth]{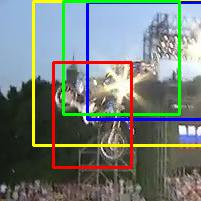}\hfill%
\includegraphics[width=\ww\textwidth]{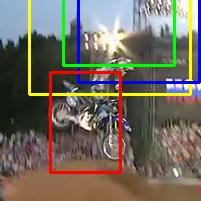}\hfill%
\includegraphics[width=\ww\textwidth]{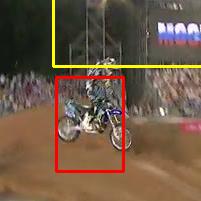}\hfill%
\includegraphics[width=\ww\textwidth]{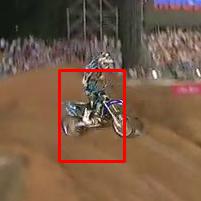}\hfill%
\\[1pt]
\includegraphics[width=\ww\textwidth]{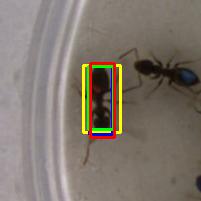}\hfill%
\includegraphics[width=\ww\textwidth]{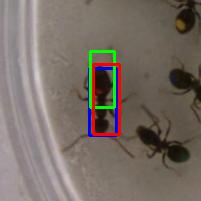}\hfill%
\includegraphics[width=\ww\textwidth]{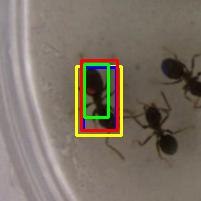}\hfill%
\includegraphics[width=\ww\textwidth]{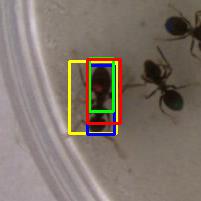}\hfill%
\includegraphics[width=\ww\textwidth]{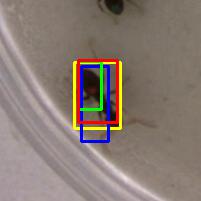}\hfill%
\includegraphics[width=\ww\textwidth]{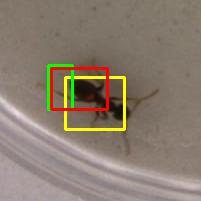}\hfill%
\includegraphics[width=\ww\textwidth]{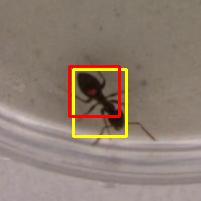}\hfill%
\includegraphics[width=\ww\textwidth]{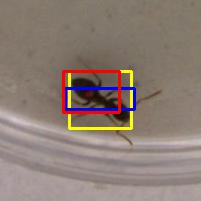}\hfill%
\includegraphics[width=\ww\textwidth]{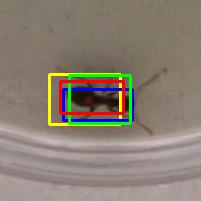}\hfill%
\includegraphics[width=\ww\textwidth]{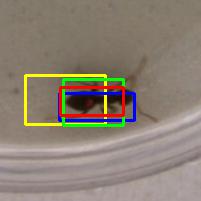}\hfill%
\\[1pt]
\includegraphics[width=\ww\textwidth]{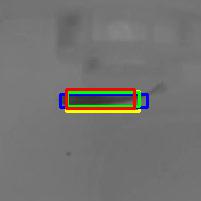}\hfill%
\includegraphics[width=\ww\textwidth]{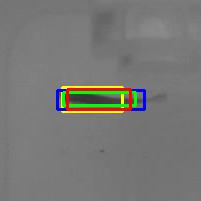}\hfill%
\includegraphics[width=\ww\textwidth]{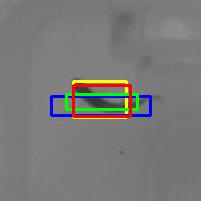}\hfill%
\includegraphics[width=\ww\textwidth]{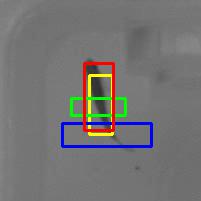}\hfill%
\includegraphics[width=\ww\textwidth]{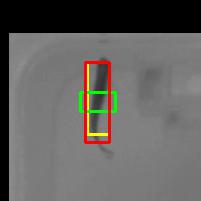}\hfill%
\includegraphics[width=\ww\textwidth]{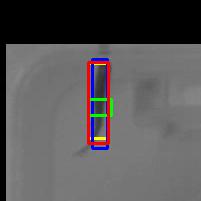}\hfill%
\includegraphics[width=\ww\textwidth]{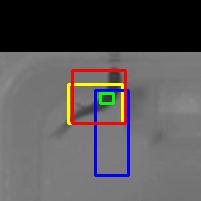}\hfill%
\includegraphics[width=\ww\textwidth]{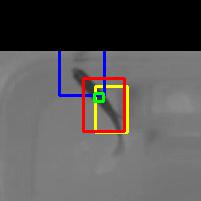}\hfill%
\includegraphics[width=\ww\textwidth]{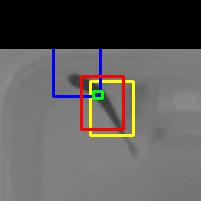}\hfill%
\includegraphics[width=\ww\textwidth]{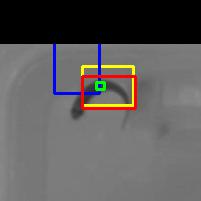}\hfill%
\caption{Visual tracking comparison between \textcolor{red}{\textbf{PBTS}} and 
the best three part-based trackers on the VOT2018 benchmark,
\textcolor{ForestGreen}{ANT}, \textcolor{blue}{DPT}, and 
\textcolor{Goldenrod}{LGT}, on the sequences \textit{motocross1}, 
\textit{ants1}, and \textit{zebrafish1} (first, second, and third rows 
respectively). Frames in which there is no bounding box for a tracker indicate
that it failed recently and will be reinitialised.}
\label{fig:PBTS_final_VOT2018_GOOD}
\end{figure*}

We performed a component ablation study using the VOT2018 benchmark to evaluate
the contribution of each key component of PBTS. 
Table~\ref{tbl:ablation_results} shows the performance of varying one 
component, keeping all others fixed. The label \textit{No alpha matting}
indicates that the procedure of \cite{init_paper} is not used and the entire
bounding box is predicted as containing the object (the default assumption of
most trackers). \textit{Uniform patch placement} indicates that the entire
patch placement scheme was not used; instead patches were placed uniformly over
the bounding box, similarly to \cite{LGT, DCCO, elastic_patches}. 
\textit{Default MBD} refers to using $b = \tfrac{1}{2}$ in 
Equation~\ref{eqn:ch3:BC}, relatively up-weighting the contribution of poorer
performing patches compared to the optimised value of $b = 1.4$. 
\textit{No local optimisation} denotes PBTS with $L=0$, restricting 
transformations to only search non-shearing, affine transformations. 
\textit{No model update} denotes setting $\beta_s = \beta_c = 0$ and using the
patch models initialised in the first frame throughout.

Table~\ref{tbl:ablation_results} shows that each component is important to the
overall performance. In particular, the initialisation via object segmentation
and placing patches at the centre of superpixels has the most influence on
tracking performance, because this avoids placing patches on relatively static
background pixels which are subsequently tracked. Using the Bhattacharyya
distance parameter $b = \tfrac{1}{2}$, gives the least reduction in
performance, but shows that the modified distance is a valuable addition.

\subsection{Qualitative Analysis}
\label{sec:qualitative_analysis}
Fig.~\ref{fig:PBTS_final_VOT2018_GOOD} shows qualitative tracking results for
PBTS compared to ANT \cite{ANT}, DPT \cite{DPT}, and LGT \cite{LGT}. In the
\textit{motocross1} sequence, only PBTS is able to cope with in-plane and 
out-of-plane rotation, as well as large flashes of light from the floodlights
(right of images). This is largely due to the enhanced PBTS patch placement
procedure, which avoids placing patches on background pixels (\eg sky), 
enabling PBTS to track the object rather than the background. This appears to
be the reason for the large expansions of the other trackers' bounding boxes in
the initial frames. 

The \textit{ants1} sequence shows PBTS coping well with rotation and a 
non-convex object, whereas both ANT and DPT fail to track the ant as it quickly 
rotates $90^\circ$. LGT also tracks the ant without failure but starts to 
drift as the ant rotates. Objects with large aspect ratios, such as the fish in
the \textit{zebrafish1}, highlight problems with trackers that fail to estimate
object rotation correctly. ANT and DPT incorrectly estimate the object's
rotation, resulting in failure for DPT and a severe miscalculation of the
object's scale by ANT.

\begin{figure}[t]
\centering
\newcommand{\ww}{0.094}%
\includegraphics[width=\ww\textwidth]{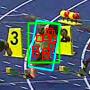}\hfill%
\includegraphics[width=\ww\textwidth]{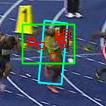}\hfill%
\includegraphics[width=\ww\textwidth]{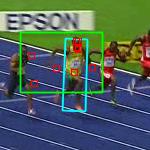}\hfill%
\includegraphics[width=\ww\textwidth]{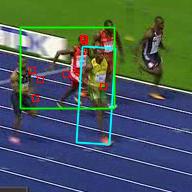}\hfill%
\includegraphics[width=\ww\textwidth]{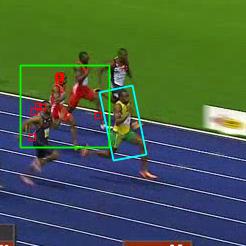}\hfill%
\\[1pt]
\includegraphics[width=\ww\textwidth]{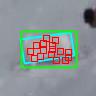}\hfill%
\includegraphics[width=\ww\textwidth]{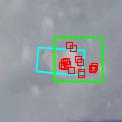}\hfill%
\includegraphics[width=\ww\textwidth]{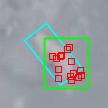}\hfill%
\includegraphics[width=\ww\textwidth]{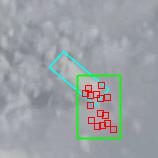}\hfill%
\includegraphics[width=\ww\textwidth]{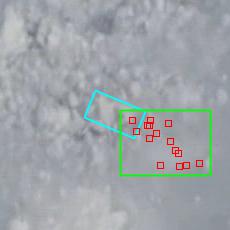}\hfill%
\\[1pt]
\includegraphics[width=\ww\textwidth]{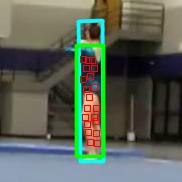}\hfill%
\includegraphics[width=\ww\textwidth]{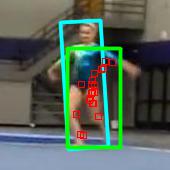}\hfill%
\includegraphics[width=\ww\textwidth]{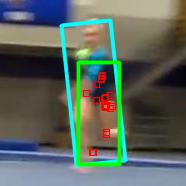}\hfill%
\includegraphics[width=\ww\textwidth]{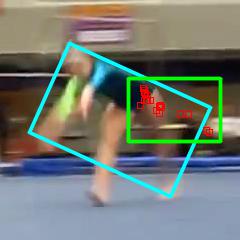}\hfill%
\includegraphics[width=\ww\textwidth]{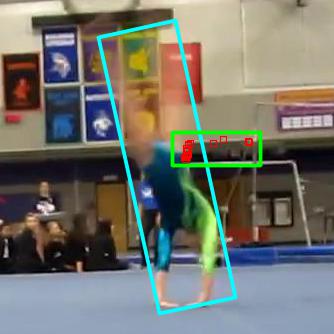}%
\caption{Failure cases in challenging VOT2018 benchmark sequences 
(\textit{bolt1}, \textit{rabbit}, and \textit{gymnastics2}). Tracked object
parts are shown in \textcolor{red}{red}, with the predicted and ground-truth
bounding boxes shown in \textcolor{ForestGreen}{green} and 
\textcolor{cyan}{cyan} respectively.}
\label{fig:PBTS_final_VOT2018_BAD}
\end{figure}
Three failure cases of the PBTS tracker are illustrated in 
Fig.~\ref{fig:PBTS_final_VOT2018_BAD}. Due to a poor segmentation in the
\textit{bolt1} video, some patches are initialised partially on the background,
which leads to those patches tracking the background, while other patches,
initialised on the runner's torso when it was in shade, match poorly as he
moves into the light and they too start to drift. In the \textit{rabbit}  
sequence the rabbit is very similar in appearance to its background, leading to
patches drifting off the rabbit as it moves across the snow. As PBTS relies
solely on a colour model it struggles to track objects that are very similar to
their background. In \textit{gymastics2} the gymnast rotates out of plane,
undergoes large amounts of deformation, has large amounts of motion blur, and 
is filmed on a hand-held, moving camera. These challenging visual attributes
combined to rapidly change the object's appearance, resulting in the tracker
underestimating the object's scale and eventually losing track of the object
completely.

\section{Conclusion}
\label{sec:conclusion}
PBTS is a novel part-based tracking framework for short-term, model-free
tracking. Essential to its performance is the patch placement mechanism that
attempts to place patches on the object being tracked, avoiding tracking the
background, and the observation that inter-frame movement is on the whole rigid
with small local deviations, which permits efficient global followed by local
search for the optimum transformation. The sparse colour model, characterising
the patch, shows surprisingly good performance for such a simple representation
and we anticipate that performance can be improved by employing \eg texture or
deep convolutional features.

{\small
\bibliographystyle{ieee}
\bibliography{refs}
}

\end{document}